\documentclass[runningheads]{llncs}
\usepackage{graphicx}
\usepackage{todonotes}
\usepackage{amsfonts}
\usepackage{amsmath}
\usepackage{wrapfig}
\usepackage{blindtext}
\usepackage{spverbatim} 
\usepackage[utf8]{inputenc}
\usepackage{tcolorbox}
\usepackage{bold-extra}

\usepackage{algorithm}
\usepackage{algpseudocode}
\begin{document}
\title{Understanding surrogate explanations: \\the interplay between complexity, fidelity and coverage}
%
\titlerunning{Understanding surrogate explanations}
%

\author{
Rafael Poyiadzi\inst{1}\and
Xavier Renard\inst{2} \and
Thibault Laugel\inst{2} \and\newline
Raul Santos-Rodriguez\inst{1} \and
Marcin Detyniecki\inst{2,3,4}
}
\authorrunning{Poyiadzi, R. et al.}
%
\institute{
University of Bristol, Bristol, United Kingdom \and
AXA, Paris, France \and
Sorbonne Université, CNRS, LIP6, F-75005, Paris, France \and
Polish Academy of Science, IBS PAN, Warsaw, Poland
}

\maketitle              
\begin{abstract}
This paper analyses the fundamental ingredients behind surrogate explanations to provide a better understanding of their inner workings. We start our exposition by considering global surrogates, describing the trade-off between complexity of the surrogate and fidelity to the black-box being modelled. We show that transitioning from global to local -- reducing coverage -- allows for more favourable conditions on the Pareto frontier of fidelity-complexity of a surrogate. We discuss the interplay between complexity, fidelity and coverage, and consider how different user needs can lead to problem formulations where these are either constraints or penalties. We also present experiments that demonstrate how the local surrogate interpretability procedure can be made interactive and lead to better explanations.

\end{abstract}

\section{Introduction}

\textit{Local surrogate explainers} is a class of \textit{post-hoc} interpretability methods whose goal is to help explain a prediction for an instance, produced by \textit{any} machine learning model. These are usually considered \textit{black-boxes}, i.e. we have no access to the model's internals; we can only observe the output for a given input. For example, let $f: \mathbb{R}^d \to \{0, 1\}$ denote a binary classifier. Then the black-box allows us to feed an instance $\boldsymbol{x}\in\mathbb{R}^d$ through $f$, and then \textit{only} observe the output, $f(\boldsymbol{x})\in\{0, 1\}$. 

The goal of local surrogates is to model the \textit{local} classification behaviour of the black-box with the instance of interest as the reference. In this context, `local' is a loose term that implies closeness with respect to a distance metric, where this distance can be geometric, e.g. Euclidean distance or weighted Manhattan distance; it can be based on semantics, e.g. all images of a certain class; or it can be any other measure that reflects closeness -- determined by the application or imposed by the user \cite{hepburn2021}.

As discussed in \cite{poyiadzi2021overlooked}, existing methods that fall in this category (such as LIME \cite{ribeiro2016should}, PALEX \cite{jia2019palex}, GSLS \cite{laugel2018b}, LEAP \cite{localembeddings}, LORE \cite{guidotti2018local}) follow a similar recipe for explaining an instance $\boldsymbol{x}$, but differ on their objectives and their understanding of locality. Practitioners and researchers alike, are now faced with a myriad of approaches with no simple way of deciding which one to use, and what the properties and assumptions of the different approaches are \cite{poyiadzi2021overlooked,kacper2019}. The common core algorithm followed by these techniques presented in \cite{poyiadzi2021overlooked} is shown in Algorithm \ref{algo:local_surrogate_recipy}. The major difference between all these approaches is how they sample the neighbourhood (line 1) and how they assign weights (line 3). These control the information to be modelled by the local surrogate. 

In this work we take a step back and try to understand local surrogates from a different point of view. We start off by discussing global surrogates and the trade-off between \textit{accuracy} and \textit{complexity}. We then introduce the notion of locality -- and, local surrogates -- as a means of narrowing down the area of interest with the aim of making the conditions of the accuracy-complexity trade-off amenable and more favourable. We consider the whole spectrum ranging from global to local and refer to this idea as \textit{coverage}. Instead of assuming the existence of a `right level of locality', we propose and study a dynamic and interactive view of locality that could lead to more information extracted from local surrogate modelling. We argue that high coverage is a favourable property of surrogates as it implies a higher level of trust, and a better understanding of generalisation of the surrogate.


\renewcommand{\algorithmicrequire}{\textbf{Input:}}
\renewcommand{\algorithmicensure}{\textbf{Output:}}

\begin{algorithm}[!t]
  \caption{Fitting an interpretable local surrogate}\label{algo:local_surrogate_recipy}
  \begin{algorithmic}[1]
      \Require{black-box $f$, instance to be explained $\boldsymbol{x}$} 
      \Ensure{Interpretable Local Surrogate}
      \State Sample, or generate, the neighbourhood $\boldsymbol{X_{N}}$ which is a means of encouraging locality to $\boldsymbol{x}$.
      \State Obtain labels for $\boldsymbol{X_{N}}$ from the black-box: $y_N = f(\boldsymbol{X_{N}})$.
      \State (Optional) Compute weights, $\boldsymbol{w_{N}}$, that measure closeness to $\boldsymbol{x}$, for all instances in $\boldsymbol{X_{N}}$.
      \State Fit a local surrogate model -from an interpretable model class- using $(\boldsymbol{X_{N}}, \boldsymbol{y_{N}}, \boldsymbol{w_{N}})$.
  \end{algorithmic}
\end{algorithm}
Our contributions are as follows:
\begin{itemize}
    \item We introduce an important property of local surrogates: coverage. Coverage controls the region of interest and determines how local or global the surrogate is.
    \item We discuss and empirically illustrate the interplay between coverage, fidelity and complexity.
    \item We propose a new interactive workflow for local surrogate modelling that depends on considering ranges of localities, rather a fixed, predetermined one.
\end{itemize}
\section{From Global to Local Surrogate Explainers}\label{sec:local_surrogates}

\subsection{Global Surrogates}

\begin{figure}[!t]
    \centering
    \includegraphics[width=\textwidth]{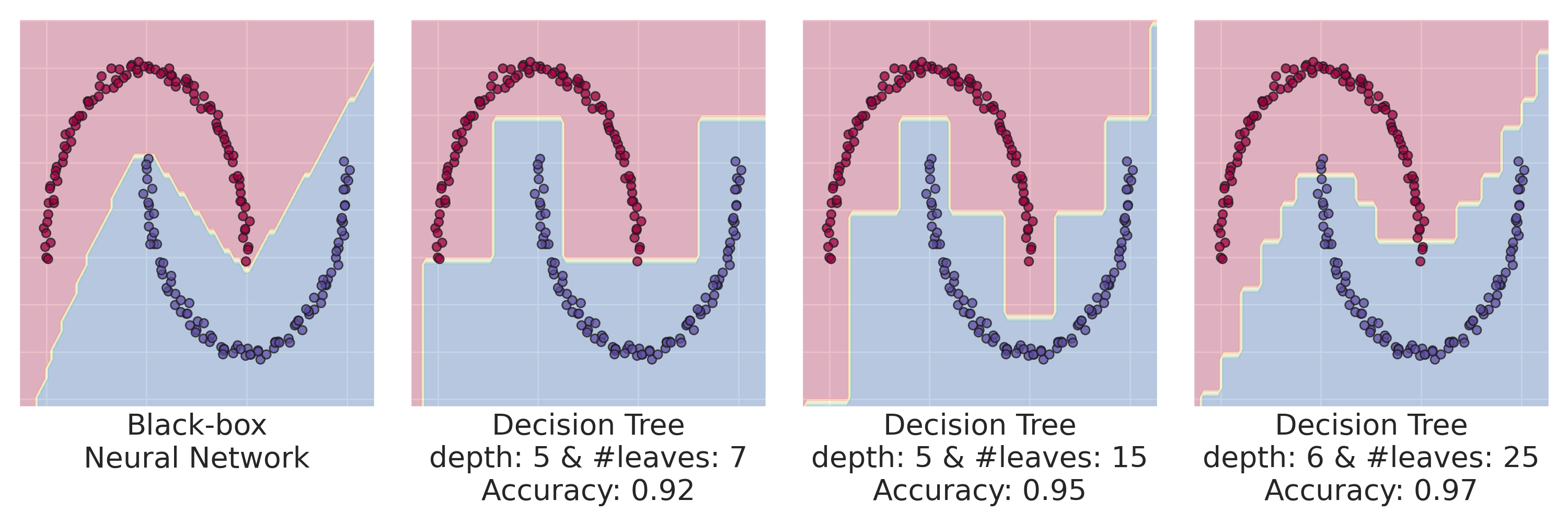}
    \caption{Toy example that illustrates the trade-off between complexity and accuracy, faced by global surrogates. Complexity, in this case, is measured by depth and number of leaves. \textbf{Accuracy is measured with respect to the black-box model, and not the original data,} $\boldsymbol{X}$. In this toy example we do this in an exhaustive manner by considering a meshgrid of points $\boldsymbol{X}_s$ spanning the range of the data, and then obtaining predictions from the black-box $f(\boldsymbol{X}_s)$. (This procedure is not recommended in general, and is only carried out for illustration purposes.)}
    \label{fig:global_surrogates_2}
\end{figure}

Surrogates refer to models fitted on top of a given initial model $f$, trained on $(\boldsymbol{X}, \boldsymbol{y})$. A surrogate $g_{s}$ would then be trained on $(\boldsymbol{X}_{s}, f(\boldsymbol{X}_{s}))$, where it could be that $\boldsymbol{X}_{s} = \boldsymbol{X}$, or not. In what follows we consider $f$ to be either a black-box or highly complex model that we wish to approximate with $g \in G$, where $G$ is a interpretable model class, such as linear models, decision trees or decision rules. 

Our assumptions on the high complexity of $f$ combined with the complexity restriction on $g$ give rise to {two, potentially conflicting, objectives: \textbf{fidelity} and \textbf{complexity}.} We want to obtain a simple interpretable surrogate $g$ that matches as close as possible the behaviour of a potentially complex model $f$.

Figure \ref{fig:global_surrogates_2} illustrates this trade-off: complexity (as measured by depth and number of leaves of a decision tree) against accuracy\footnotemark. On the other hand, accuracy is computed with respect to the black-box model, and not to the original data ($\boldsymbol{X}$) only. In this toy example, we measure accuracy in an exhaustive manner by considering a meshgrid of points $\boldsymbol{X}_s$ spanning the range of the data, and then obtaining predictions from the black-box $f(\boldsymbol{X}_s)$. 
\footnotetext{In this work we use fidelity and accuracy interchangeably to refer to metrics that measure how well the local surrogate models the behaviour of the black-box. The metrics we consider are: (1) the proportion of correct predictions, and (2) true positive rate and true negative rate.}

\begin{figure}[!t]
    \centering
    \includegraphics[width=0.45\textwidth]{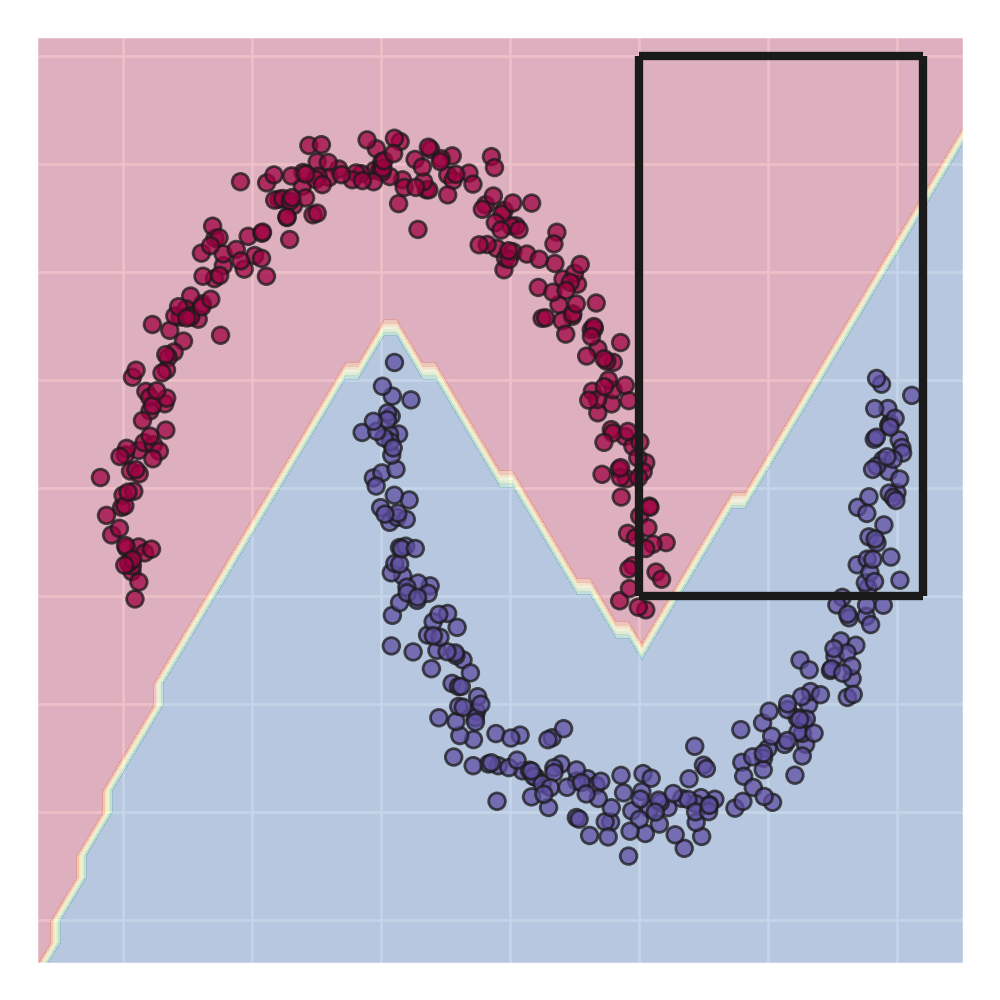}
    \caption{A black-box's decision boundary to illustrate how focusing on a smaller region  makes modelling easier. Approximating the black-box model in its entirety would be challenging for a linear model, but focusing on just the right region makes it possible to achieve a high level of fidelity with the desired level of complexity.}
    \label{fig:focus_locally}
\end{figure}
          
{But, what if we do not care about being accurate everywhere? If we \textit{only} care about a specific instance, can we do better than a global surrogate?} The motivation is that if instead of trying to model the black-box's behaviour \textit{everywhere}, we only model it regionally, then the task would require a surrogate model with a lower complexity. This is exemplified in Figure \ref{fig:focus_locally}. Approximating the black-box model in its entirety would be challenging for a linear model, but focusing on just the right region makes it possible to achieve a high level of fidelity, with the desired level of complexity. This is what local surrogate explainers aim to do. A key problem is \textit{how} to define the relevant region, and then proceed to model it.

\subsection{Local Surrogate Explainers}

\begin{figure}[!t]
    \centering
    \includegraphics[width=0.55\textwidth]{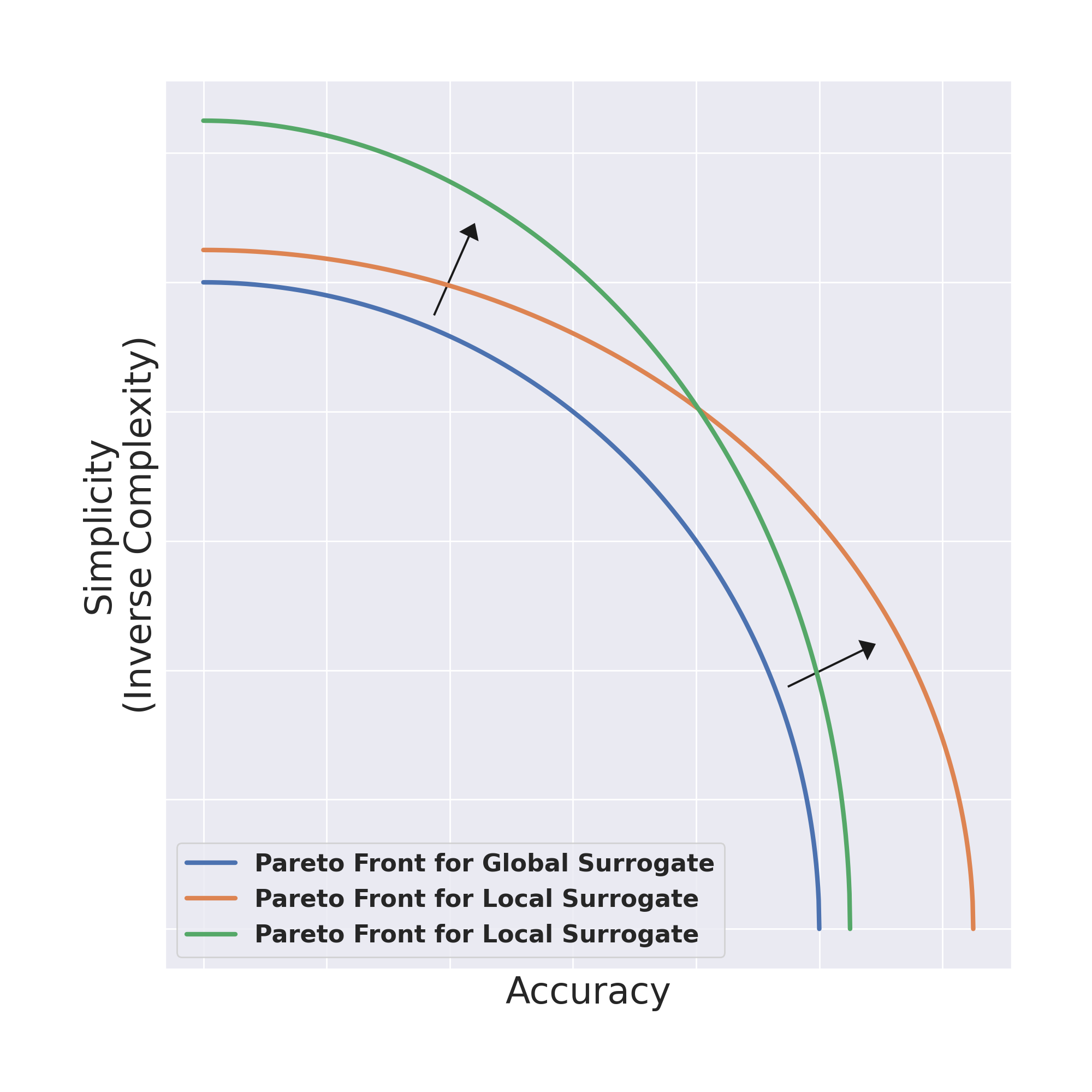}
    \caption{Pareto frontiers for different types of surrogates. This illustrates how the Pareto frontier for a global surrogate (blue) can be made more favourable by only focusing locally (green and orange).}
    \label{fig:pareto_fronts}
\end{figure}


Local surrogate explainers, as compared to global surrogate explainers, (implicitly) tackle the accuracy-complexity trade-off by first relaxing the objective of accuracy: from requiring the approximation to be accurate everywhere, to requiring it to be accurate \textit{locally} only. By reducing the focus from global to local -- as shown, for example, in Figure \ref{fig:focus_locally} -- we can shift the Pareto frontier further out, making conditions more favourable, as shown in Figure \ref{fig:pareto_fronts}. This Figure illustrates the advantages of reducing the region of interest for the surrogate, and considering a local, rather than a global model. The blue line presents a Pareto frontier for a global surrogate, for example the one presented in Figure \ref{fig:global_surrogates_2}. Green and oranges lines show Pareto frontiers for two local surrogates, that highlight that when moving from a global model to a local one, better accuracy can be achieved under the same complexity, or, alternatively, a model with lower complexity can be obtained with a similar accuracy score. For example, consider the second sub-plot from Figure \ref{fig:global_surrogates_2} that shows the decision boundary of a decision tree with \texttt{depth = 5} and \texttt{number of leaves = 7} achieving an accuracy of $0.92$. Then, a decision tree fitted on the enclosed region in Figure \ref{fig:focus_locally} with \texttt{depth = 5} and \texttt{number of leaves = 7} could achieve an accuracy of $\geq 0.92$, or that for a desired accuracy of $0.92$, a decision tree with less complexity would be needed. This, of course, does not come for free - reducing the area of focus comes with its own implications. Consider shifting from the global case to a very small region, such that you can fit an accurate model that is also interpretable. How useful would it be? Probably not much. This illustrates the need for the third objective: \textbf{coverage}. The preceding argument does not imply that more coverage is necessarily better, although it might as well be. For the same accuracy and complexity, one should be prefer a model with a higher coverage as it would be more general.
By moving from global coverage to partial/local coverage, we improve the Pareto frontier on accuracy and complexity. 

To summarise, we identify the following properties of interest for local surrogates that we will further analyse empirically. 

    \begin{enumerate}
        \item \texttt{\textbf{Complexity (interpretability)}.} While this depends on the application and the user, a few generally accepted model classes are decision trees, linear models and rule-based models. These are not always human-interpretable. A deep decision tree, or a linear model with many non-zero entries, can be difficult to interpret and offer little to no information. Therefore, complexity constraints might need to be considered.
        \item \texttt{\textbf{Fidelity (accuracy)}.} The fitted surrogate model should have as high fidelity as possible with respect to the black-box model.
        \item \texttt{\textbf{Coverage (region of interest)}.} In general, high coverage is preferable to low coverage, as it would make the explanations easier to generalise and more trustworthy.
    \end{enumerate}
The final objectives for local surrogate explainers will depend on these three ingredients under different formulations. Example scenarios could be as follows:

\begin{enumerate}
    \item For a fixed coverage, obtain the most accurate linear model with at most $k$ non-zero coefficients.
    \item For an accuracy that is greater than $x\%$, understand the interaction of coverage and complexity.
    \item For a linear model with at most $k$ non-zero coefficients, understand the interaction of coverage and accuracy.
\end{enumerate}

In the next section we present experiments to illustrate the practical interest of considering a wide range of localities and also show the interplay of the three properties -- coverage, fidelity and complexity. Finally, we build upon these to show how fitting multiple surrogate models -- instead of just one -- can lead to more informed decisions and extract meaningful interpretations. This is related to the field of uncertainty estimation in interpretable machine learning. In \cite{slack2020much} the authors estimate Credible Intervals (CIs) that capture the uncertainty associated with each feature importance in local explanations. They then use the CIs to inform sampling; with regards to the size of the sample and the location of the sampling. In \cite{gosiewska2019ibreakdown} authors use bootstrapping to approximate the distribution of variable contributions in local explanations, with the goal of understanding model level uncertainty. In our case, we demonstrate how bootstrapping \cite{bootstrap} can be used to provide uncertainty estimates for the learned models to enhance understanding.
\section{Experiments}\label{sec:experiments}

\begin{figure}[!t]
    \centering
    \includegraphics[width=\textwidth]{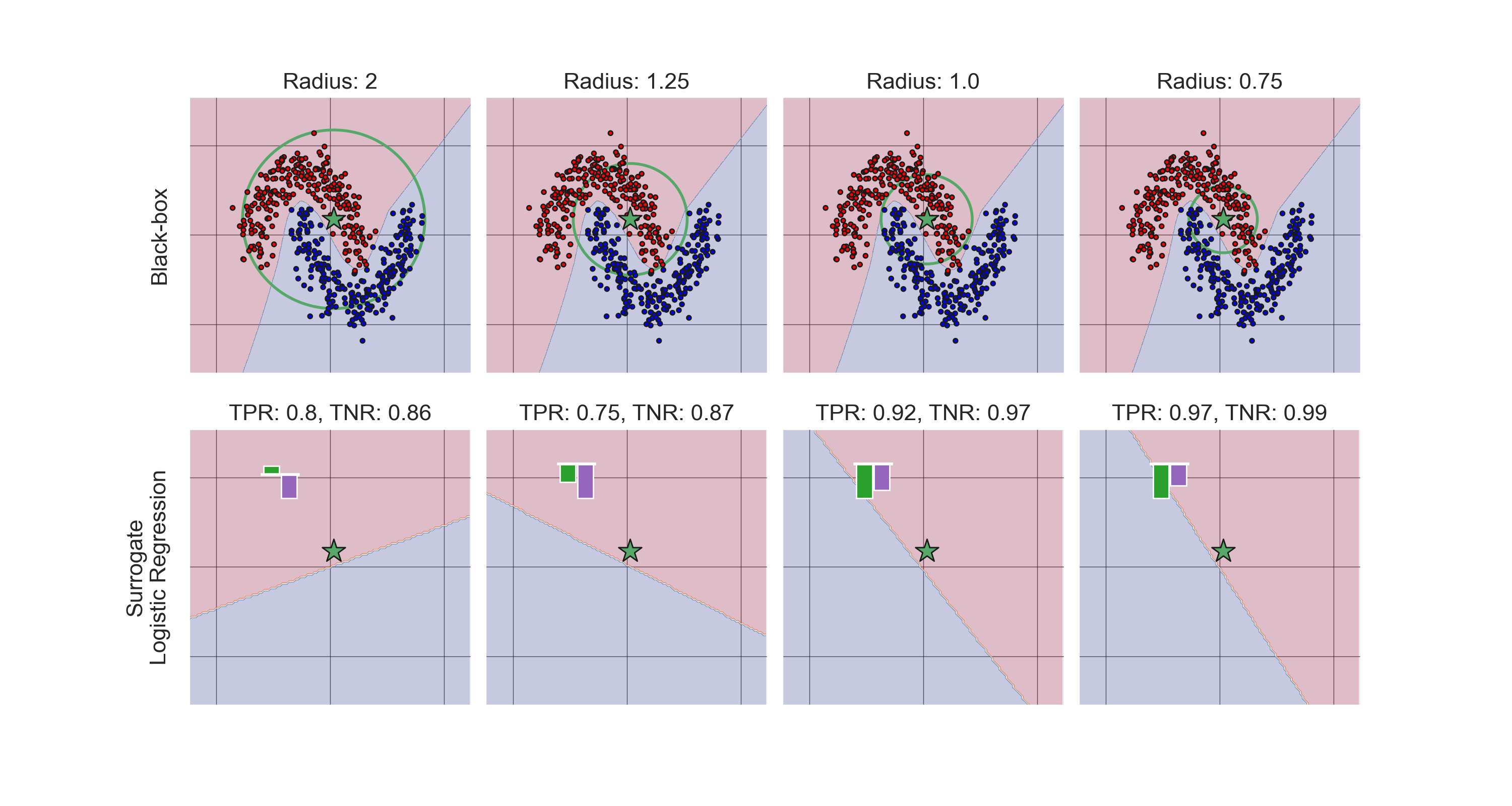}
    \caption{Surrogate models fitted on neighbourhoods generated from a range of radii. \textbf{top row:} decision boundary of a black-box (a neural network) that has been trained on the half-moons data. \textbf{bottom row:} decision boundaries of the surrogates (logistic regression). The small embedded bar-plot presents the coefficients of the two features. On the titles we note the \textit{true-positive rate} (TPR) and the \textit{true negative rate} (TNR) as measures of fidelity to the black-box. In all cases we plot the instance to be explained (green star) and the hyper-spheres (green circles) from which we sample the neighbourhood uniformly at random.}
    \label{fig:local_surrogates_lr}
\end{figure}
In this section we present experiments on toy datasets to illustrate the advantages of viewing locality as a continuum rather than keeping it fixed. In all that follows we model coverage by introducing a hyper-sphere, of radius $r$, centered on $\boldsymbol{x}$ -- the instance to be explained. The neighbourhood is then generated by sampling uniformly at random from within this hyper-sphere. Unless specified otherwise, the size of the neighbourhood is fixed at $2000$ points. 


In Figure \ref{fig:local_surrogates_lr} we consider a range of radii and the surrogates obtained by training on the corresponding neighbourhoods. On the top row we see the decision boundary of a black-box (a neural network) that has been trained on the \texttt{half-moons} dataset. We also see the instance to be explained (green star) and the hyper-spheres (green circles) from which we sample the neighbourhoods. On the second row we have the decision boundaries of the surrogates (logistic regression) and the small embedded bar-plot presents the coefficients of the two features. On every sub-figure on the second row we also note the \textit{true-positive rate} (TPR) and the \textit{true negative rate} (TNR) as measures of fidelity to the black-box.

We see that the \texttt{radius} -- which controls the coverage -- has significant impact on both the accuracy metrics and the coefficients of the surrogate models learned. In the far-left pair of sub-figures -- which is the most global -- of the four, the coefficient corresponding to the first feature is positive, while then it turns negative. We also see an increase in the accuracy metrics when moving from the first case to the second, and when moving from the second to third. When comparing the last two we see that coefficients and accuracy have almost settled, and the choice would depend on the user.

\begin{figure}
    \centering
    \includegraphics[width=\textwidth]{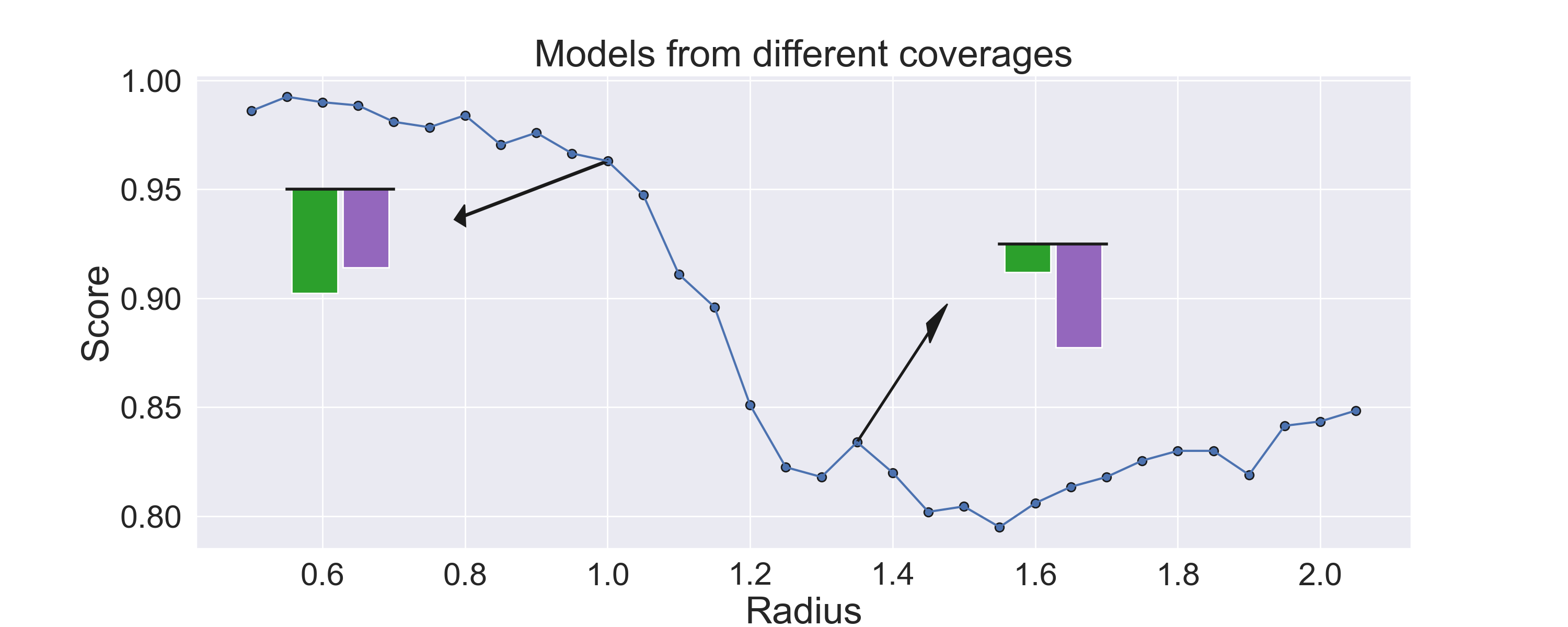}
    \caption{Surrogate models fitted on neighbourhoods generated from a range of radii. We show radius on the x-axis, and accuracy on the y-axis. The two embedded bar-plots correspond to the coefficients of the surrogate models (logistic regression) trained on neighbourhoods with radius as shown by the arrows.}
    \label{fig:local_surrogates_different_coverage_lr}
\end{figure}

In Figure \ref{fig:local_surrogates_different_coverage_lr} we consider a finer spectrum of radii. This gives the bigger picture of how models trained on different levels of coverage perform (with regards to accuracy) and what their learned parameters imply for the instance to be explained.

\begin{figure}[!t]
    \centering
    \includegraphics[width=\textwidth]{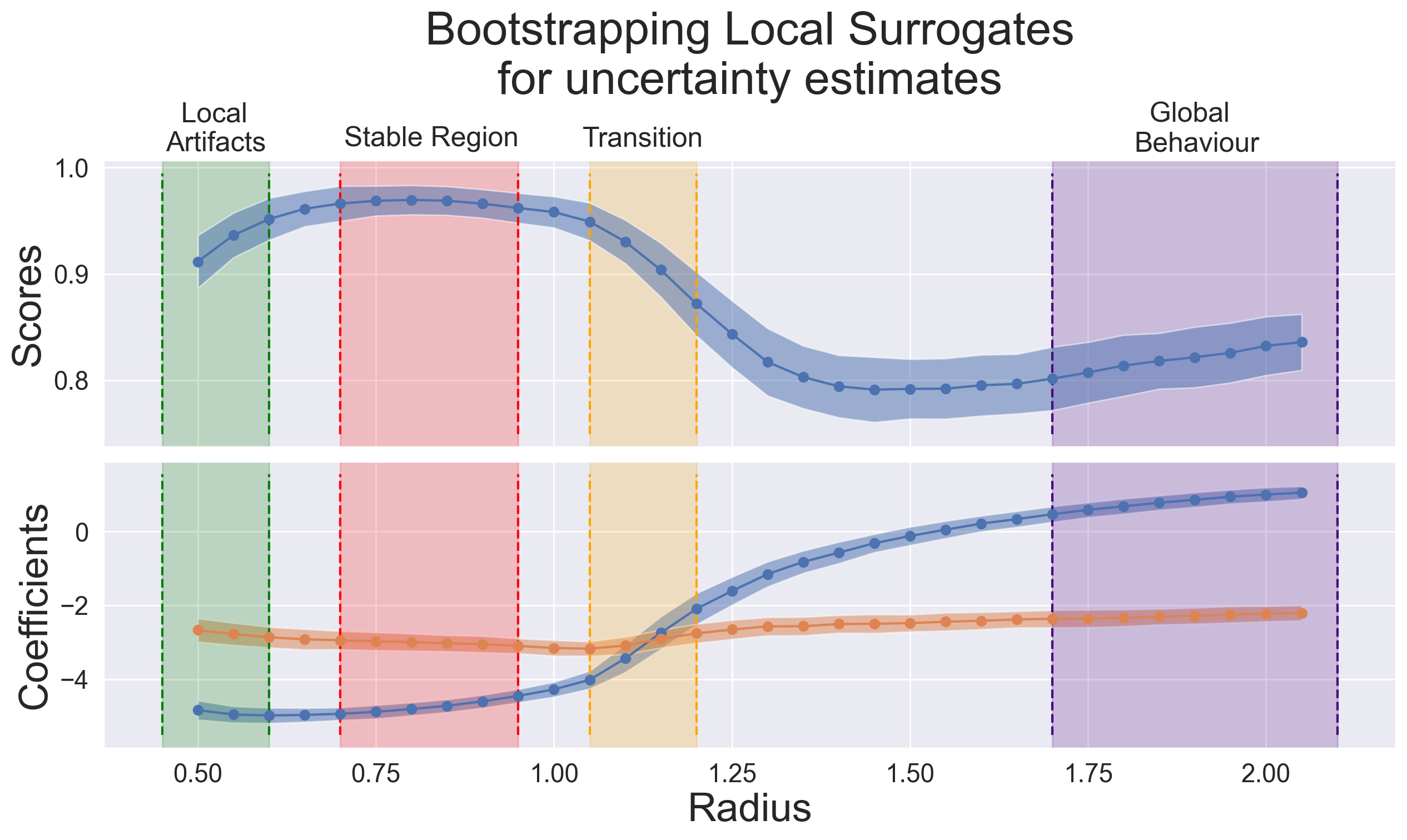}
    \caption{Bootstrapping the local surrogate. For every value of \texttt{radius} we train $500$ local surrogates on different draws of a neighbourhood. In this case we consider neighbourhoods of size $200$ to gain a better sense of variability (as opposed to neighbourhoods of size $2000$ as in other experiments). \textbf{top:} scores of bootstrapped surrogate models. \textbf{bottom:} coefficients of bootstrapped surrogate models. In both cases we plot the mean $\pm$1std. Vertical shaded regions are manually annotated to denote different behaviours of bootstrapped models and how they could be interpreted. }
    \label{fig:bootstrap_local_surrogates_lr}
\end{figure}

In Figure \ref{fig:bootstrap_local_surrogates_lr} we introduce uncertainty in the explanations produced (for the same instance) to enhance the understanding of the behaviour of the models through the use of bootstrapping \cite{bootstrap}. We highlight four regions in the plots that were manually annotated accordingly. As we vary the coverage of (bootstrapped) local surrogates, we see that both the accuracy score and the coefficients change. We observe a general trend of decreasing accuracy (top sub-figure) as the radius increases, as well as a small increase in the standard deviation between the bootstrapped surrogate models. We also observe how the coefficients of the models vary with the radius, and how one of them transitions from negative to positive. 

\begin{figure}[!ht]
    \centering
    \includegraphics[width=\textwidth]{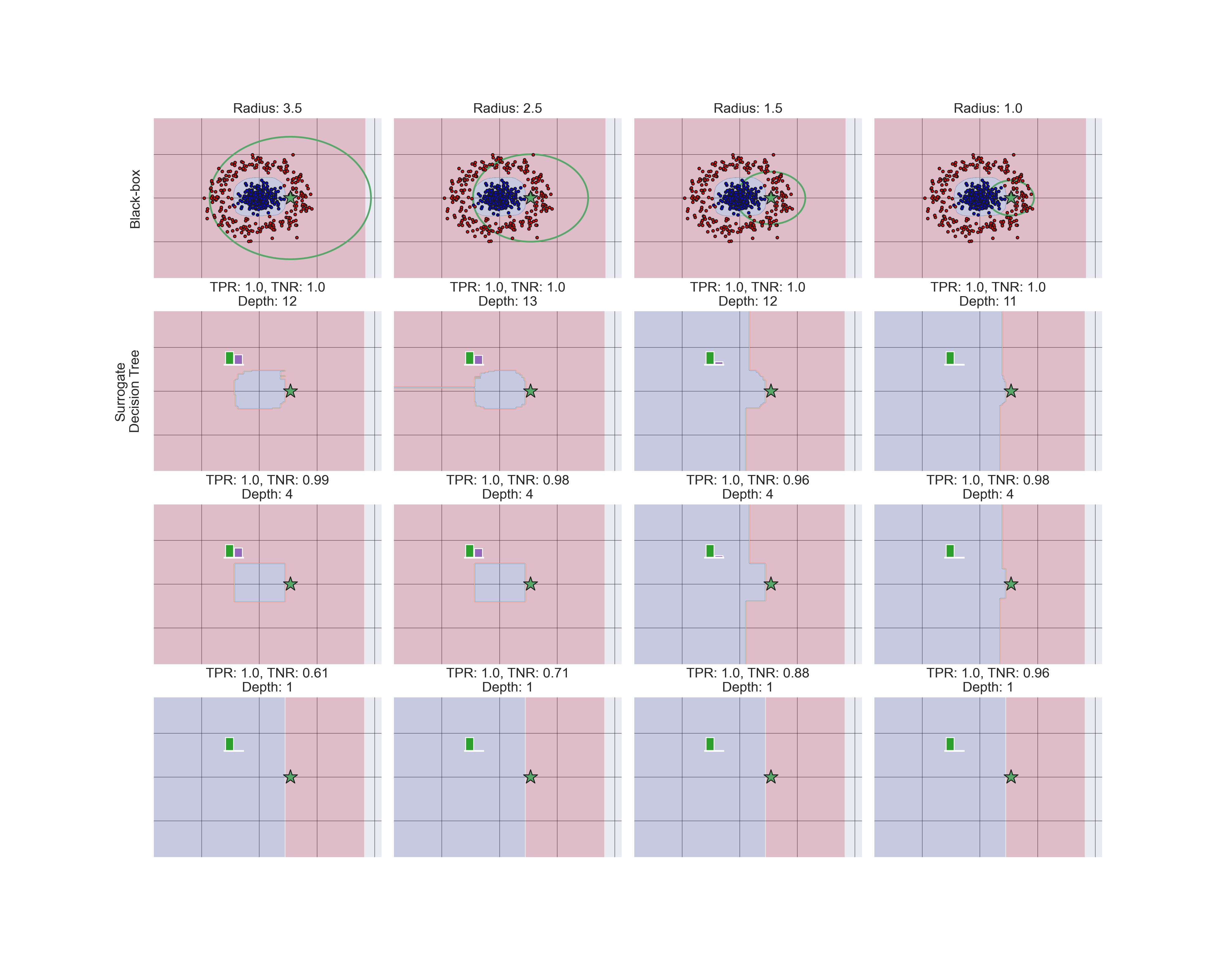}
    \caption{Surrogate models fitted on neighbourhoods generated from a range of radii. \textbf{top row:} decision boundary of a black-box (a neural network) that has been trained on the \texttt{circles data}. \textbf{second row:} decision boundaries of the surrogates (decision tree). \textbf{third \& fourth rows:} decision boundaries of the surrogates but this time with \texttt{max\_depth} fixed to $4$ and $1$ respectively. On the titles we note the \textit{true-positive rate} (TPR) and the \textit{true negative rate} (TNR) as measures of fidelity to the black-box. In all cases we plot the instance to be explained (green star) and the hyper-spheres (green circles) from which we sample the neighbourhood. The small embedded bar-plot presents the feature importance measures extracted from the surrogate}
    \label{fig:local_surrogate_dtc}
\end{figure}

In Figure \ref{fig:local_surrogate_dtc} we let our interpretable model class be decision trees. We now explore the setting where we constraint complexity of the surrogate model, and observe the model's accuracy. In the second row of the figure we see a succession of models trained on neighbourhoods with decreasing radius. With no complexity constraints the models achieve good fidelity scores and we also observe a decrease in complexity with a decrease in coverage. In the last two rows of the figure we impose complexity constraints on the training of the model: on the third row we restrict models to have at most $depth=4$ and on the last row we restrict models to have at most $depth=1$.

\begin{figure}[!t]
    \centering
    \includegraphics[width=\textwidth]{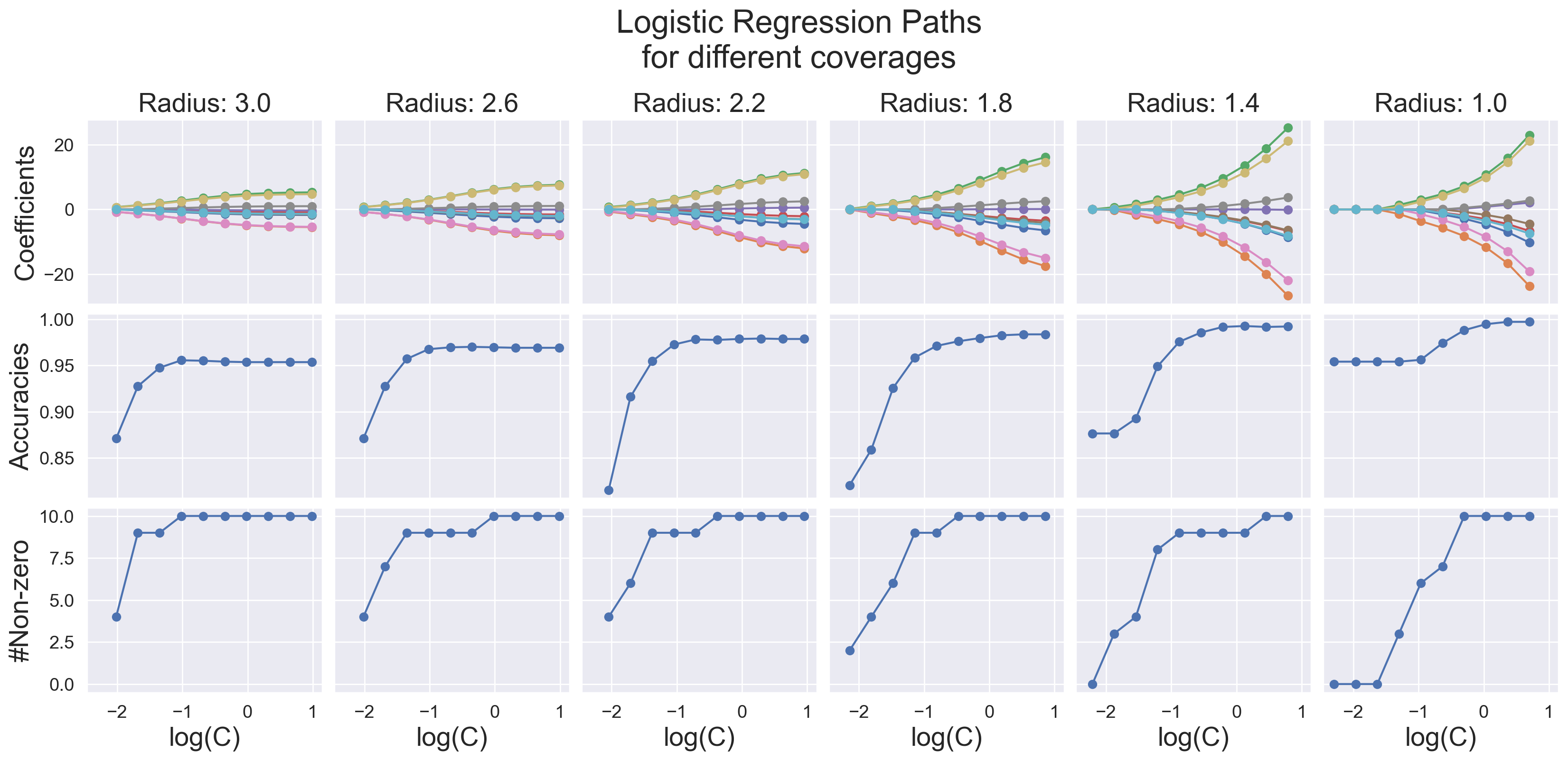}
    \caption{Lasso paths together with the corresponding accuracy and $l_0$-norm of solution. For this we use the \texttt{diabetes} dataset and transform it to a classification task by thresholding on the median value of the target variable. \textbf{top row:} regularisation paths for lasso logistic regression. Several models are trained on a range of decreasing regularisation parameters. On the left-end, regularisation ($C$) is strongest -- and all coefficients are zero -- while on the right-end regularisation is at its weakest. Each colour corresponds to the `path' of a different coefficient. \textbf{middle row:} accuracy scores for the models along the path. \textbf{bottom row:} number of non-zero coefficients for the models along the path. \textbf{columns:} each column corresponds to models trained on neighbourhoods generated from a different radius -- which is specified at the top of each column.}
    \label{fig:lasso_paths}
\end{figure}

In Figure \ref{fig:bootstrap_local_surrogates_lr} we kept the complexity fixed (number of non-zero coefficients) and experimented with the interaction of coverage and accuracy. In problems on higher-dimensions we can also experiment with number of non-zero coefficients. A popular formulation for linear models is the Lasso, where the introduced $l_1$-norm penalty leads to sparser solutions. It is used in core machine learning and statistics, and in interpretable machine learning for this reason. In Figure \ref{fig:lasso_paths} we experiment with the \texttt{diabetes} dataset. Our model class is logistic regression, but now with $l_1$-norm penalty. A useful tool for analysis and visualisation for this model class is the so-called `regularisation path' of multiple solutions to the lasso formulation. A solution is obtained for successively relaxed penalty parameters - reducing the effect of the sparsity inducing $l_1$-norm penalty. Then for every solution you can observe how the coefficients vector varies. Examples of such paths are shown in the first row of Figure \ref{fig:lasso_paths}, where every colour corresponds to one coefficient and shows how it varies along the `path' of decreasing regularisation strength ($C$). In the same figure for a range of radii we plot the lasso path (top), accuracy scores of the models (middle), and the number of non-zero coefficients of the solutions (bottom). We observe a general trend of the accuracy scores path shifting upwards, as the radius decreases, and we also observe the beginning of the complexity paths (bottom) to shift downwards. This follows along the lines of previous observations where as we decrease the coverage, we can obtain a models with non-decreasing accuracy and non-increasing complexity.

\newpage
\section{Discussion \& Future Work}\label{sec:discussion}
In this paper we presented an alternative approach for understanding and informing the construction of local surrogates. We have discussed and shown the interaction between coverage, fidelity and complexity. We have proposed a workflow for interpretable local surrogates which is interactive and where we consider a range of localities instead of a fixed, predetermined locality.

Future work will involve further exploration of the different problems and limitations posed in the paper, namely:
\begin{itemize}
    \item\texttt{\textbf{Coverage:}} We have introduced coverage as an important property of local surrogates. We model coverage as the radius of a hyper-sphere centered on the instance to be explained (from which we sample the neighbourhood). This is easy to understand and implement, but also very generic. However, one could argue whether this is the best way of defining coverage. Also, ideally we will want to design procedures for increasing/decreasing coverage that can adapt to the user's needs. For example, a straightforward extension to the hyper-sphere could be a hyper-ellipsoid where radii along different dimensions are derived from the user's needs, or preferences. 
\item\texttt{\textbf{Sampling:}} Here we  used a simple sampling regime where we sampled the neighbourhood uniformly at random from the hyper-sphere. We also did not make use of weights during training. However, sampling can change to make training more efficient, allowing training the surrogate with fewer training samples. Also, classification tasks with high-dimensionality are particularly challenging as they would require a larger size for a neighbourhood, rather than a fixed size.

\item \texttt{\textbf{Uncertainty:}} Uncertainty is important in interpretability as it signals trust to the output of the model; be it a predictive model, or be it a local surrogate explainer. In this work we have presented the use of bootstrapping the surrogate by training on multiple (different) draws of the neighbourhood. This is of course computationally expensive. There might be better, computationally efficient methods for deriving uncertainty estimates for local surrogates. For Generalised Linear Models (GLM) there exist ways of deriving such uncertainty estimates without having to resort to multiple runs, but these would depend on assumptions to hold true.
\end{itemize}
\section*{Acknowledgements}
This work was partially funded by the UKRI Turing AI Fellowship EP/V024817/1.
\clearpage
\newpage
\bibliographystyle{splncs04}
\bibliography{bibliography}

\end{document}